\documentclass[sigconf]{acmart}  

\usepackage{subcaption}
\usepackage{multirow} 
\usepackage{arydshln} 
\usepackage{enumitem} 
\usepackage{framed}
\usepackage{fancybox}

\newcommand{\dialogueboxtitle}[1]{%
    {\bfseries\small #1}\\[-0.3em]%
    \noindent\rule{\linewidth}{0.4pt}\vspace{0.3em}%
}

\newenvironment{dialoguebox}[2][]{%
    \def\dialogueframetitle{#2}%
    \setlength{\OuterFrameSep}{4pt}%
    \setlength{\FrameSep}{7pt}%
    \colorlet{shadecolor}{gray!10}%
    \begin{shaded}%
    \dialogueboxtitle{\dialogueframetitle}%
    \small
}{%
    \end{shaded}%
}

\definecolor{dialoguebg}{RGB}{248,248,250}
\definecolor{dialogueframe}{RGB}{130,130,140}
\definecolor{dialoguetitle}{RGB}{60,60,70}

\newenvironment{dialogueboxE}[2][0.9\linewidth]
{%
  \begin{center}
  \setlength{\fboxrule}{0.6pt}%
  \setlength{\fboxsep}{8pt}%
  \begin{Sbox}%
  \begin{minipage}{\dimexpr#1-2\fboxsep-2\fboxrule\relax}%
  {\color{dialoguetitle}\bfseries #2}\par
  \vspace{0.45em}
  \hrule height 0.4pt
  \vspace{0.55em}
}
{%
  \end{minipage}%
  \end{Sbox}%
  \fcolorbox{dialogueframe}{dialoguebg}{\TheSbox}%
  \end{center}
}

\newlist{dialoguelist}{itemize}{1}
\setlist[dialoguelist,1]{%
    label={},
    leftmargin=0.2em,
    labelsep=0pt,
    align=left,
    itemsep=0.3em,
    parsep=0pt,
    before=\small\setlength{\parskip}{0pt},
    after=\setlength{\parskip}{0pt}
}

\newenvironment{coloredbox}[4][0.95\linewidth]
{%
  \def\cbwidth{#1}%
  \def\cbframecolor{#2}%
  \def\cbbgcolor{#3}%
  \def\cbtitle{#4}%
  \begin{center}
  \setlength{\fboxrule}{0.6pt}%
  \setlength{\fboxsep}{8pt}%
  \begin{Sbox}%
  \begin{minipage}{\dimexpr\cbwidth-2\fboxsep-2\fboxrule\relax}%
  {\color{\cbframecolor}\bfseries \cbtitle}\par
  \vspace{0.45em}
  \hrule height 0.4pt
  \vspace{0.55em}
}
{%
  \end{minipage}%
  \end{Sbox}%
  \fcolorbox{\cbframecolor}{\cbbgcolor}{\TheSbox}%
  \end{center}
}

\definecolor{observerbg}{RGB}{245,250,245}
\definecolor{observerframe}{RGB}{70,120,70}

\definecolor{drawerbg}{RGB}{245,248,252}
\definecolor{drawerframe}{RGB}{70,100,145}

\definecolor{variousbg}{RGB}{252,250,240}
\definecolor{variousframe}{RGB}{150,125,50}

\definecolor{reasonerbg}{RGB}{252,245,245}
\definecolor{reasonerframe}{RGB}{145,75,75}

\definecolor{judgebg}{RGB}{248,248,250}
\definecolor{judgeframe}{RGB}{110,110,120}

\newenvironment{observer}[2][0.95\linewidth]
  {\begin{coloredbox}[#1]{observerframe}{observerbg}{#2}}
  {\end{coloredbox}}

\newenvironment{drawer}[2][0.95\linewidth]
  {\begin{coloredbox}[#1]{drawerframe}{drawerbg}{#2}}
  {\end{coloredbox}}

\newenvironment{various}[2][0.95\linewidth]
  {\begin{coloredbox}[#1]{variousframe}{variousbg}{#2}}
  {\end{coloredbox}}

\newenvironment{reasoner}[2][0.95\linewidth]
  {\begin{coloredbox}[#1]{reasonerframe}{reasonerbg}{#2}}
  {\end{coloredbox}}

\newenvironment{judge}[2][0.95\linewidth]
  {\begin{coloredbox}[#1]{judgeframe}{judgebg}{#2}}
  {\end{coloredbox}}

\usepackage{pifont}

\newcommand{\legendsquarehex}[1]{%
  \textcolor[HTML]{#1}{\rule{2ex}{2ex}}%
}

\AtBeginDocument{%
  }

\setcopyright{acmlicensed}
\copyrightyear{2026}
\acmYear{2026}
\acmDOI{XXXXXXX.XXXXXXX}
\acmConference[]{}{}

\makeatletter
\def\city#1{\global\@ACM@citypresenttrue}
\def\country#1{\global\@ACM@countrypresenttrue}
\makeatother

\renewcommand\footnotetextcopyrightpermission[1]{}
\begin{document}

\title{Using Machine Mental Imagery for Representing Common Ground in Situated Dialogue}


\author{Biswesh Mohapatra}
\authornote{Both authors contributed equally to this work. Names are listed in alphabetical order.}
\affiliation{%
  \institution{Inria}
  \city{Paris}
  \country{France}}
\email{biswesh.mohapatra@inria.fr}

\author{Giovanni Duca}
\authornotemark[1]
\authornote{Work done during Master's thesis at Inria.}
\affiliation{%
  \institution{University of Trento}
  \city{Paris}
  \country{France}}
\email{giovanni.duca-1@studenti.unitn.it}

\author{Laurent Romary}
\affiliation{%
  \institution{Inria}
  \city{Paris}
  \country{France}}
\email{laurent.romary@inria.fr}

\author{Justine Cassell}
\affiliation{%
  \institution{Inria, Carnegie Mellon University}
  \city{Paris}
  \country{France}
  }
\email{justine.cassell@inria.fr}

\renewcommand{\shortauthors}{Mohapatra et al.}

\begin{abstract}
Situated dialogue requires speakers to maintain a reliable representation of shared context rather than reasoning only over isolated utterances. Current conversational agents often struggle with this requirement, especially when the common ground must be preserved beyond the immediate context window. In such settings, fine-grained distinctions are frequently compressed into purely textual representations, leading to a critical failure mode we call \emph{representational blur}, in which similar but distinct entities collapse into interchangeable descriptions. This semantic flattening creates an illusion of grounding, where agents appear locally coherent but fail to track shared context persistently over time. Inspired by the role of mental imagery in human reasoning, and based on the increased availability of multimodal models, we explore whether conversational agents can be given an analogous ability to construct some depictive intermediate representations during dialogue to address these limitations. Thus, we introduce an active visual scaffolding framework that incrementally converts dialogue state into a persistent visual history that can later be retrieved for grounded response generation. Evaluation on the IndiRef benchmark shows that incremental externalization itself improves over full-dialog reasoning, while visual scaffolding provides additional gains by reducing representational blur and enforcing concrete scene commitments. At the same time, textual representations remain advantageous for non-depictable information, and a hybrid multimodal setting yields the best overall performance. Together, these findings suggest that conversational agents benefit from an explicitly multimodal representation of common ground that integrates depictive and propositional information.
\end{abstract}

%

\begin{teaserfigure}
    \centering
    \includegraphics[width=\textwidth]{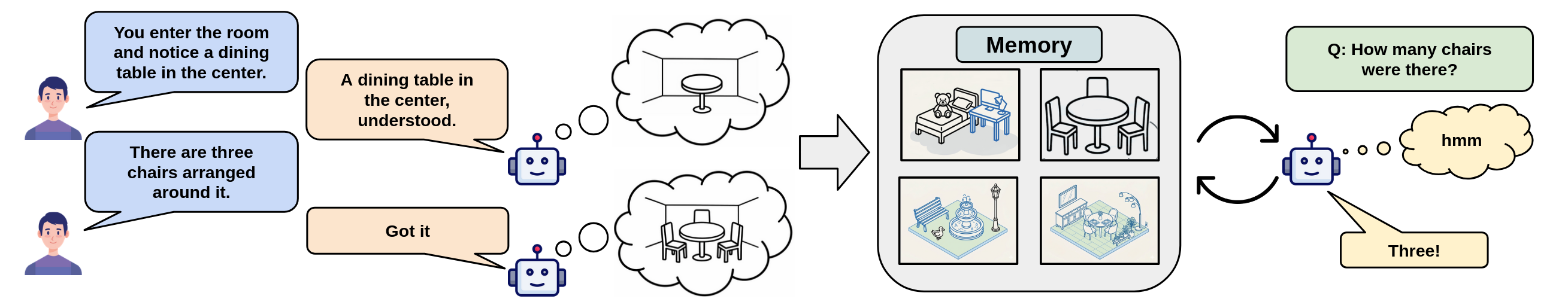}
    \caption{As the human describes the scene, the agent incrementally constructs and updates a mental image of the established information. It can be later utilized to refer back to the common ground established.}
    \Description{Proper description of the image, not a caption alongside it but what is going on. probably ALT_TEXT? not sure}
    \label{fig:mental_imagery_teaser}
\end{teaserfigure}


\maketitle


\section{Introduction}

Human communication often involves a trade-off between minimizing communicative effort and maintaining sufficient clarity for successful understanding~\cite{kempRegier2012kinship, zipf1949human}. To maximize efficiency, speakers routinely rely on underspecification, conveying only part of the required information and leaving the remainder to be recovered from shared context and the physical environment, or requests for clarification from the interlocutor \cite{pezzelleDealingSemanticUnderspecification2023, clark1991grounding}. Communication therefore succeeds not simply because utterances are informative in isolation, but because interlocutors continuously coordinate on how those utterances should be interpreted in context. This coordination process, through which participants establish, maintain, and repair a shared understanding over the course of interaction, is known as \emph{conversational grounding} \cite{Clark1989ContributingTD, clark1991grounding}. In situated dialogues \cite{kelleher-kruijff-2006-incremental, kontogiorgos2019uncertainty, narayan-chen-etal-2019-collaborative}, grounding becomes particularly demanding because speakers must align not only on linguistic content, but also on entities, events, spatio-temporal relations, and perspective-dependent references while the shared context itself evolves over time. Grounding in such settings is therefore not merely a matter of local individual utterances, but of dynamically constructing and maintaining an evolving representation of the shared physical and linguistic context~(common ground) over extended exchanges that can support later reasoning, disambiguation, and reference~\cite{huaTalkLessInteract2024, Mohapatra2024ConversationalGA}. For conversational agents, this requirement is especially important in environments where relevant information must remain accessible even after it has fallen outside the immediate context window.

While human interlocutors can resolve these semantic gaps using non-linguistic context, nonverbal means such as eye gaze~\cite{nakano-etal-2003-towards}, and an evolving model of the shared environment, contemporary conversational agents remain largely constrained to text and the context window. Consequently, they often struggle with the kind of naturally efficient, underspecified language that human dialogue routinely relies on, leaning instead toward exhaustive and token-heavy interpretations \cite{pezzelleDealingSemanticUnderspecification2023, otani-etal-2023-underspecification}. Prior work has shown that, while Large Language Models (LLMs -- which have become the backbone of modern conversational agents) can exhibit locally appropriate grounding behaviors such as providing acknowledgments \cite{mohapatra-etal-2024-evaluating}, they often fail to later \emph{use} the established common ground reliably. More importantly, this failure points to a deeper representational bottleneck: when common ground is stored in purely textual forms, such as summaries or ontological structures, it remains vulnerable to compression, semantic drift, and ambiguity, and often fails to preserve the fine-grained distinctions required for robust long-horizon grounding \cite{mohapatra2026frame}. 

This motivates our central hypothesis that conversational agents may benefit from access to a more depictive representational substrate, a hypothesis made possible by the increasing availability of multimodal models~\cite{openai2024gpt4ocard, bai2025qwen3vl}. When people describe a scene, event, or spatial layout, the listener often forms and incrementally updates a mental image of what is being described as can be seen in Figure~\ref{fig:mental_imagery_teaser}. Human interlocutors do not rely on propositional language alone when reasoning about such scenes; they construct situation models of the described world and may utilize perceptual simulation during comprehension, which helps preserve visuospatial distinctions and supports the resolution of referential ambiguity \citep{Zwaan1998SituationMI, perceptual_symbol_systems, Fincher-Kiefer2001}. Inspired by this, our work explores whether a conversational model can be given a similar ability to \emph{construct and update mental imagery during conversation} as a depictive intermediate representation that supports both immediate understanding and later conversational use, and that is realized by multimodal models.

We therefore frame this paper as an exploration of \textbf{\textit{Visual Scaffolding}} for conversational grounding. By Visual Scaffolding, we mean the use of generated visual artifacts as an external memory substrate that incrementally materializes the agent’s current interpretation of an unfolding dialogue and supports later querying of the evolving shared state. We investigate whether a model can actively construct and update such depictive artifacts over time, using \textbf{\textit{Style-as-Semantics}} to externalize its evolving interpretation of the scene. These artifacts serve as verifiable intermediate commitments. They force the system to resolve at least part of the underspecified content of the dialogue into a concrete structure that can later be reasoned over. 

Our study is thus guided by the following research question: Can the implicit, evolving state of a situated dialogue be translated into a consistent sequence of explicit visual artifacts that functions as a verifiable history of conversational interpretation?
To address this, we introduce a framework that externalizes dialogue state tracking into dynamic visual memory. Rather than passively accumulating text, the system incrementally transforms the evolving dialogue into a sequence of discrete visual sketches. During inference, an agentic multimodal Retrieval Augmented Generation (RAG) pipeline queries this repository to support grounded response generation. 

To distinguish the benefits of visual externalization from the more general benefits of incremental state construction, we compare this visual condition with a textual ablation that constructs dense scene summaries instead of images while keeping everything else the same.

We evaluate this framework on the Meetup subset of the IndiRef dataset using an LLM-as-a-Judge evaluation~\cite{mohapatra2026frame}, a setting that is particularly well suited to studying persistent grounding. We instantiate this framework using multiple State-of-the-Art~(SOTA) Vision-Language Models (VLMs) capable of multimodal reasoning and iterative image editing. The results show that incrementally externalizing common ground improves performance over direct full-dialog reasoning, indicating that the organization of conversational evidence into discrete, content-addressable states is itself a source of improvement. Within this agentic framework, the two modalities exhibit complementary strengths. Visual Scaffolding is particularly effective at preserving fine-grained perceptual distinctions over long interactions, while explicit text remains better suited to representing the non-depictable aspects of the shared state. Results from the hybrid setting further suggest that visual and textual artifacts do not merely provide redundant views of the same information, but instead preserve partially non-overlapping evidence about the evolving common ground. These findings broadly indicate that robust conversational grounding may benefit from an incremental memory where agents maintain both depictive and propositional representations of shared context using separate modalities. This work has direct implications for multimodal dialogue systems that must reason over shared physical environments.

\section{Related Work}

\paragraph{\textbf{State Tracking in Situated Dialogue.}} 
Maintaining a persistent representation of Common Ground over extended time periods remains a primary challenge in situated dialogue \cite{clark1991grounding, udagawaMaintainingCommonGround2021}.
Traditional Dialogue State Tracking (DST) relies on slot-value architectures, framing state maintenance as the population of predefined schemas \cite{jacqminYouFollowMe2022b, kontogiorgos2019uncertainty}.
While highly interpretable, this formulation is poorly suited to continuous spatial environments, where ad hoc perceptual attributes and emergent spatial relations cannot be neatly mapped to static, \textit{a priori} slots. To address this inflexibility, recent approaches utilize dynamic Knowledge Graphs (dynamic KGs) to model context as an evolving network of entities and relationships \cite{bu-etal-2025-query}.
However, graph-based parsers face significant hurdles in spoken dialogue. Because human interlocutors frequently use underspecified referring expressions, resolving these shifting references into a coherent symbolic graph without generating duplicate nodes or erroneous links has been a challenge. Consequently, modern architectures largely favor utilizing RAG~\cite{lewis2020RAG} to manage dialogue history.
However, applying standard text-based RAG to situated environments restricts the agent to propositional representations. In the process, important details about the physical state can be lost, making retrieval harder when successful reasoning depends on spatial layout, perspective, or fine-grained scene distinctions\cite{benderStochasticParrots2021, mohapatra2026frame}.

\paragraph{\textbf{Cognitive Offloading.}} 
When a shared environment becomes too complex to keep track of mentally, people often draw or sketch to reduce cognitive effort \cite{fanDrawingVersatileCognitive2023, tylen2020evolution}. In line with Dual-Coding Theory \cite{paivioDualCodingTheory1991}, these external visualizations help connect abstract language to concrete spatial structure, making relationships easier to represent and reason about \cite{hueyVisualExplanationsPrioritize2023, Vinker2022CLIPasso, fanPragmaticInferenceVisual2020}. By contrast, contemporary conversational agents remain largely passive.
Even while powered by Vision-Language Models (VLMs) capable of ingesting rich multimodal inputs, these agents do not autonomously generate visual artifacts to manage their computational load or resolve informational ambiguity \cite{huaTalkLessInteract2024, pezzelleDealingSemanticUnderspecification2023}.
This human capacity to externalize the cognitive load motivates our framework: we argue that to better prevent representational blur, situated agents can actively synthesize their common ground into verifiable multimodal states.

\paragraph{\textbf{Visual Scaffolding.}} 
Recent efforts to integrate visual scaffolding into compositional multimodal reasoning fall into two primary paradigms: visual attention and latent scaffolding.
Attention-based methods enforce semantic grounding by directing the model’s focus to relevant parts of a fixed visual scene \cite{Shtedritski2023CLIPRedCircle}.
More recent approaches go a step further by actively selecting and tracking important entities across multiple turns of interaction \cite{jianLargeLanguageModels2024, liuTakingNotesBrings2025}. However, these tracking techniques remain fundamentally constrained by the input: if the necessary spatial information is absent or ambiguous within the original pixels, the system cannot process it \cite{fu2024blink}.
Alternatively, latent scaffolding bypasses explicit rendering by having models form internal visual representations within their hidden states~\cite{liImagineReasoningSpace2025, yangMachineMentalImagery2025}.
While computationally efficient, these representations cannot be directly inspected, making it difficult to tell whether the model is reasoning about space in a grounded way or simply producing plausible outputs~\cite{chengVisualThoughtsUnified2025}.
Our framework diverges from past approaches by prioritizing explicit generation, allowing us to inspect what the model has inferred, preserved, and grounded over time.
Further, synthesizing intermediate visual states has been shown to successfully resolve semantic ambiguity prior to agent action in Robotics \cite{niGenerateSubgoalImages2024}.
We extend this principle to tracking the established common ground.
By translating the dialogue history into a persistent sequence of schematic sketches, we construct a verifiable record of the interaction over time.
This explicit rendering forces the model to commit to concrete spatial layouts, preventing further hallucinations~\cite{yang3dmem3DScene2025}.

\section{Task and Datasets} \label{sec:task_datasets}

To validate the hypothesis that explicit visual scaffolding is possible, and that it enhances situated reasoning, we require an environment that necessitates collaborative spatial coordination. We thus utilize the test cases derived from the MeetUp! corpus~(MU!)~\citep{ilinykh2019meetup} in the IndiRef benchmark \citep{mohapatra2026frame} for interaction traces and evaluation.


\paragraph{\textbf{The MeetUp! Corpus}}
\textit{MU!} provides real-time, text-based interactions between two people engaged in a cooperative coordination game. The environment is modeled as a discrete 2D grid of connected rooms, where each node is visualized using a real-world image from the ADE20k dataset \citep{zhou2017sceneParsingADE20k}. 
The task enforces two critical constraints that necessitate active state tracking: \textit{partial observability} and \textit{symmetric visibility}. Agents  only see their immediate room and available paths; neither possesses a global map. Consequently, both interlocutors must actively negotiate meaning and construct a shared spatial understanding through dialogue to successfully navigate to a common target destination. This constraint penalizes superficial understanding, requiring agents to describe rooms, maintain shared knowledge over time, track the partner’s perspective, and resolve ambiguities, all of which are central challenges in conversational grounding.

\begin{figure}[!hbtp]
    \centering
    \includegraphics[width=\linewidth]{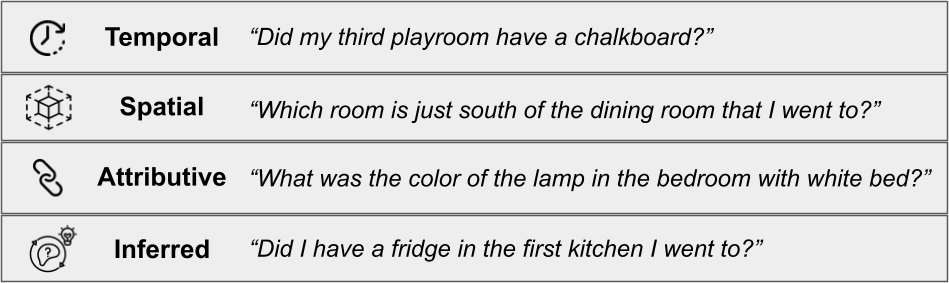}
    \caption{IndiRef question examples.}
    \label{fig:indiref_example}
\end{figure}

\paragraph{\textbf{The IndiRef Benchmark}}
To evaluate the established common ground, we utilize the IndiRef benchmark which augments the \textit{MU!} transcripts with granular Question-Answer (QA) pairs designed to probe specific referential failures.
The benchmark evaluates agents across four semantic dimensions: 
\textbf{Temporal} (relying on chronological event sequences), 
\textbf{Spatial} (relying on spatial reasoning), 
\textbf{Attributive} (is-a or has-a relations), and 
\textbf{Inferred} (relying on unstated logical inferences, such as ``Does your bathroom also have yellow tiles?'' Implies my bathroom has yellow tiles). Examples of each are provided in Figure \ref{fig:indiref_example}. Crucially, IndiRef introduces a strict perspective-taking requirement. Questions frequently employ deictic markers (e.g., ``\textit{my car}'' vs ``\textit{your car}"), forcing the agent to map pronouns to the correct speaker's visual trajectory. This explicit separation of ``Self" versus ``Other" further provides a way to validate the perspective separation ability of our system.

\section{Methodology} \label{sec:methodology}

\begin{figure*}[!htbp]
    \centering
    \includegraphics[width=0.9\linewidth]{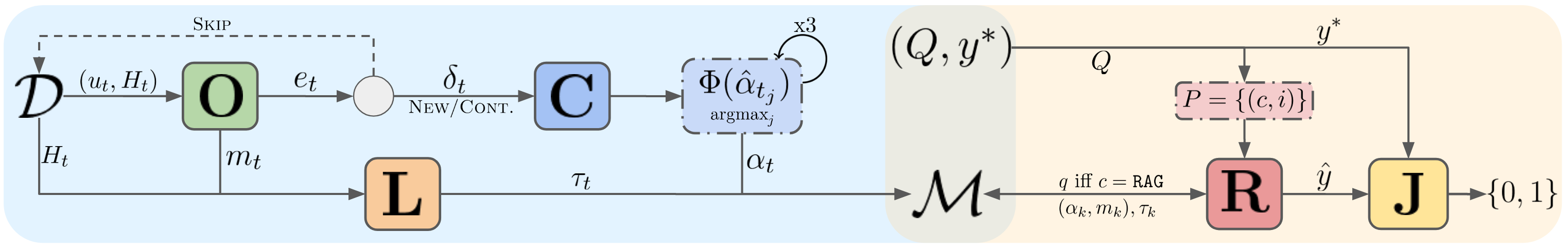}
\caption{Two-phase pipeline. \textit{Phase 1} (left): the Observer, 
Constructor, and Linker incrementally externalize the dialogue 
$\mathcal{D}$ into artifacts $\alpha_t$ and cross-scene triplets $\tau_t$, 
stored in the memory bank $\mathcal{M}$. \textit{Phase 2} (right): given 
question $Q$, the Reasoner produces a plan $\mathcal{P} = \{(c, i)\}$ 
and selectively retrieves from $\mathcal{M}$ to produce $\hat{y}$, 
evaluated by the Judge against $y^*$.}
        \label{fig:OCR}
\end{figure*}

We instantiate the pipeline with Qwen-family models (see §\ref{sec:implementation}). We propose a two-phase framework for \emph{incrementally externalizing common ground} in situated dialogue~(see Figure~\ref{fig:OCR}). In the first phase, the evolving interaction is incrementally transformed into a sequence of explicit artifacts that represent the perceived common ground being constructed over time. In the second phase, these artifacts are used as a retrieval source for grounded inference.

Let $\mathcal{D} = \{u_1^x, \dots, u_T^x\}$ denote a situated dialogue, where $u_t^x$ is an utterance produced by speaker $x$ at turn $t$. Our goal is to transform $\mathcal{D}$ into an augmented memory bank $\mathcal{M} = \{\alpha_1, \dots, \alpha_K\}$, where each artifact $\alpha_k$ is an explicit representation of the dialogue state established up to a particular point in the interaction. These artifacts are intended as intermediate commitments about the evolving common ground. To isolate the effect of representational modality, we instantiate $\alpha_k$ differently across two conditions using the same incremental segmentation and retrieval pipeline:
\begin{itemize}
    \item \textbf{Visual Condition:} $\alpha_k = \mathcal{I}_k$, where $\mathcal{I}_k$ is an explicitly generated schematic visual artifact.
    \item \textbf{Textual Condition:} $\alpha_k = \mathcal{S}_k$, where $\mathcal{S}_k$ is a dense propositional summary of the same scene-level state.
\end{itemize}

IndiRef involves two parts: a questioner, who asks about the shared environment, and an answerer, who utilizes the information established in the common ground to answer it. Because grounding is speaker-relative, we maintain two separate artifact sequences using the answerer’s perspective. One sequence stores what the questioner has established (according to the answerer), and the other stores what the answerer has established (according to the answerer). At query time, the Reasoner answers by taking the role of the answerer.

\subsection{Incrementally Externalizing Dialogue State}

The transformation from $\mathcal{D}$ to $\mathcal{M}$ is governed by three modules: an \textit{Observer}, which determines when the shared state changes; a \textit{Constructor}, which externalizes that state into an explicit artifact $\alpha$; and a \textit{Linker}, which records relations across artifacts that are difficult to preserve within a single local representation.

\textbf{The Observer (O)}: The first challenge is to determine which parts of the dialogue should update the established common ground. Not every utterance warrants artifact construction since many utterances can leave the shared common ground unchanged. The \textit{Observer} therefore decides whether the current turn establishes a new $\alpha$, updates the current $\alpha$, or should be skipped. Formally, at each turn $t$, the Observer examines the current utterance $u_t$ together with a contextual history $H_t$ and produces a discrete edit action $e_t$, a scene descriptor $\delta_t$, and auxiliary metadata $m_t$:

\vspace{-3mm}
\begin{equation}
    e_t, \delta_t, m_t \sim \pi_{obs}(u_t, H_t)
    \label{eq:observer}
\end{equation}
\vspace{-3mm}

where $e_t \in \{\textsc{New}, \textsc{Continue}, \textsc{Skip}\}$. A $\textsc{New}$ action is used when the utterance signals a transition to a new local scene. Each $\textsc{New}$ action produces a new image or summary, allowing temporal information to accumulate across the sequence. A $\textsc{Continue}$ action is used when the utterance adds, removes, or revises grounded information about the active scene. A $\textsc{Skip}$ action is used when no artifact update is warranted. Importantly, $H_t$ is constructed as a visually segmented context window containing the utterances linked to the active frame and its immediately preceding frame. To mark discontinuity explicitly, we insert a boundary token at the scene transition \texttt{<scene\_change>}. 

The Observer also factorizes each update into two complementary outputs. The scene descriptor $\delta_t$ contains the visually depictable content to be externalized into an artifact, while $m_t$ stores non-depictable but still relevant information, such as explicit negations, intentions, discourse-level qualifications, or topological movement cues. This separation is important because common ground in situated dialogue is only partially depictable: some constraints are better preserved visually, while others must remain propositional.

\textbf{The Constructor (C)}: Whenever $e_t \in \{\textsc{New}, \textsc{Continue}\}$, the \textit{Constructor} materializes the current interpretation of the scene into an explicit artifact. In the visual condition, this artifact is a schematic image $\mathcal{I}_t$; in the textual condition, it is a dense scene summary $\mathcal{S}_t$. The role of the Constructor is therefore not merely generative. It operationalizes the core hypothesis of the paper by forcing the system to \emph{externalize} its current interpretation of common ground into a persistent, inspectable intermediate representation.


To reduce hallucinations, we use a generate-verify-select procedure. Let $G(\cdot)$ denote the artifact generator and $V(\cdot)$ a verifier. The scene descriptor $\delta_t$ is first converted into a set of atomic facts $\mathcal{F} = \{f_1, \dots, f_m\}$ for verification. We then generate $J$ candidate artifacts $\hat{\alpha}_{t_j}$ conditioned on $\delta_t$ and the previous scene state $\alpha_{t-1}$:
\begin{equation}
    \hat{\alpha}_{t_j} \sim G(\delta_t, \mathcal{\alpha}_{t-1}), \quad j \in \{1, \dots, J\}
\end{equation}


Each candidate is scored by a faithfulness function $\Phi$, defined as the proportion of verified atomic facts that it satisfies:
\begin{equation}
    \Phi(\hat{\alpha}_{t_j}) = \frac{1}{|\mathcal{F}|} \sum_{i=1}^{|\mathcal{F}|} \mathbb{I}[V(\hat{\alpha}_{t_j}, f_i)]
\end{equation}
The stored artifact is then selected as
\begin{equation}
    \alpha_t = \operatorname*{argmax}_{\hat{\alpha}_{t_j}} \Phi(\hat{\alpha}_{t_j})
\end{equation}

\begin{figure*}[!htbp]
    \centering
    \resizebox{0.9\linewidth}{!}{%
    \begin{minipage}[t]{0.51\textwidth}
        \vspace{0pt}
        \begin{dialoguebox}[width=\linewidth]{Verbatim Trace (ID: 129\_383\_68\_12)}
            \begin{dialoguelist}
                \item \textbf{[Turn 26] B}: Im in a home office \hfill ($\textsc{New} \rightarrow \texttt{B\_3}$)
                \item \textbf{[Turn 27] A}: ok let me move around \hfill ($\textsc{Skip}$)
                \item \textbf{[Turn 28] B}: It also has a drum set in it \hfill ($\textsc{Continue}$)
                \item \textbf{[Turn 29] B}: 2 guitars on the wall \hfill ($\textsc{Continue}$)
                \item \textbf{[Turn 30] B}: I moved north from a kitchen to get here \hfill ($\tau \text{: } \texttt{B\_3 north\_of B\_2}$)
            \end{dialoguelist}
        \end{dialoguebox}
    \end{minipage}\hspace{0.02\textwidth}
    \begin{minipage}[t]{0.45\textwidth}
        \vspace{0pt}
        \captionsetup[subfigure]{font=scriptsize}
        \begin{subfigure}[t]{0.32\linewidth}
            \centering
            \includegraphics[width=\linewidth]{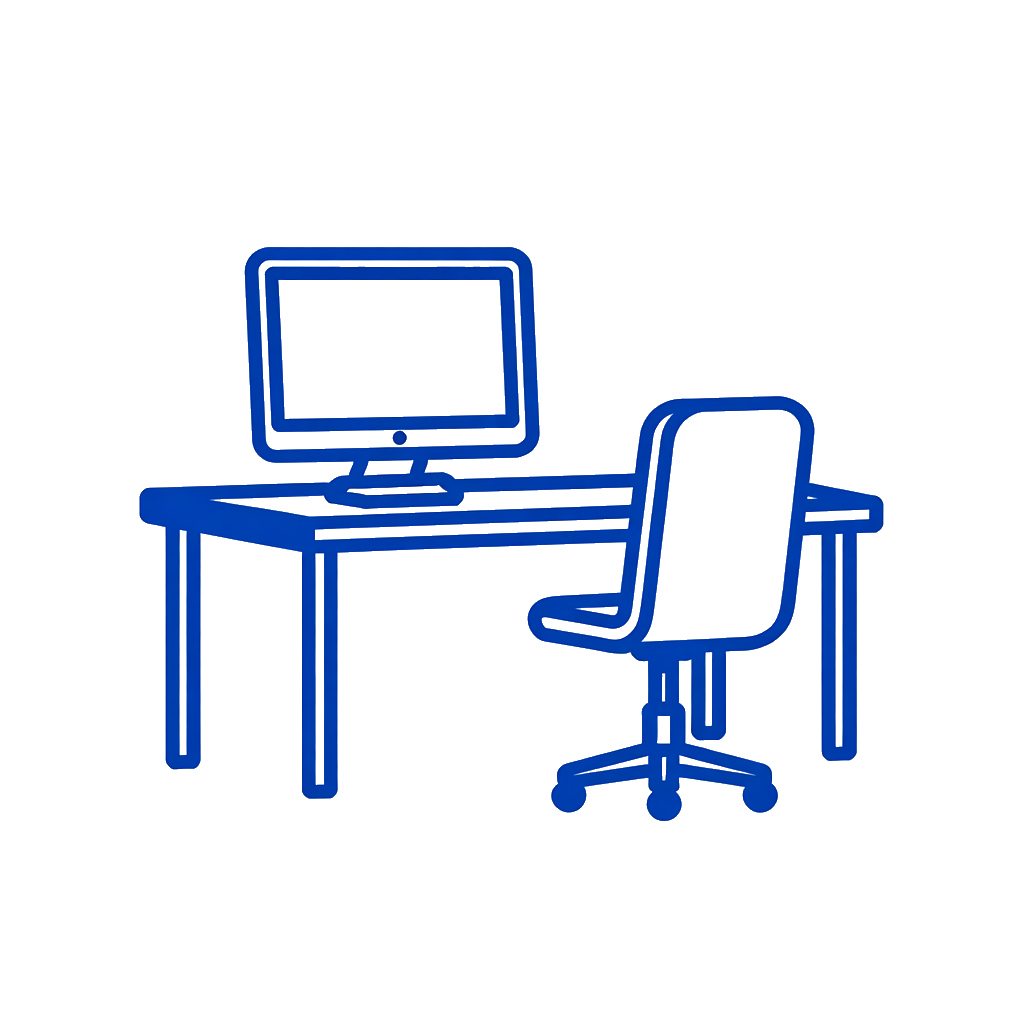}
            \caption{T26: Office created.}
        \end{subfigure}\hfill
        \begin{subfigure}[t]{0.32\linewidth}
            \centering
            \includegraphics[width=\linewidth]{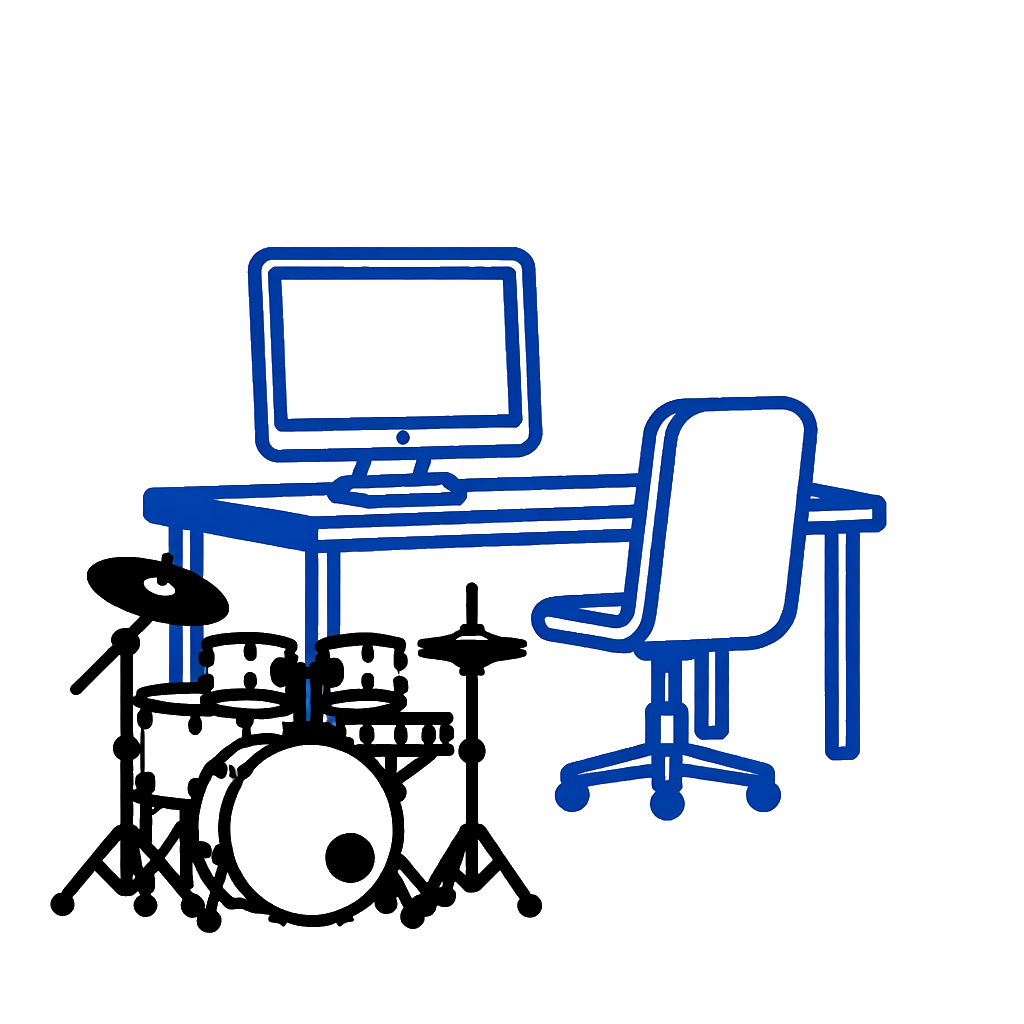}
            \caption{T28: Drum set added.}
        \end{subfigure}\hfill
        \begin{subfigure}[t]{0.32\linewidth}
            \centering
            \includegraphics[width=\linewidth]{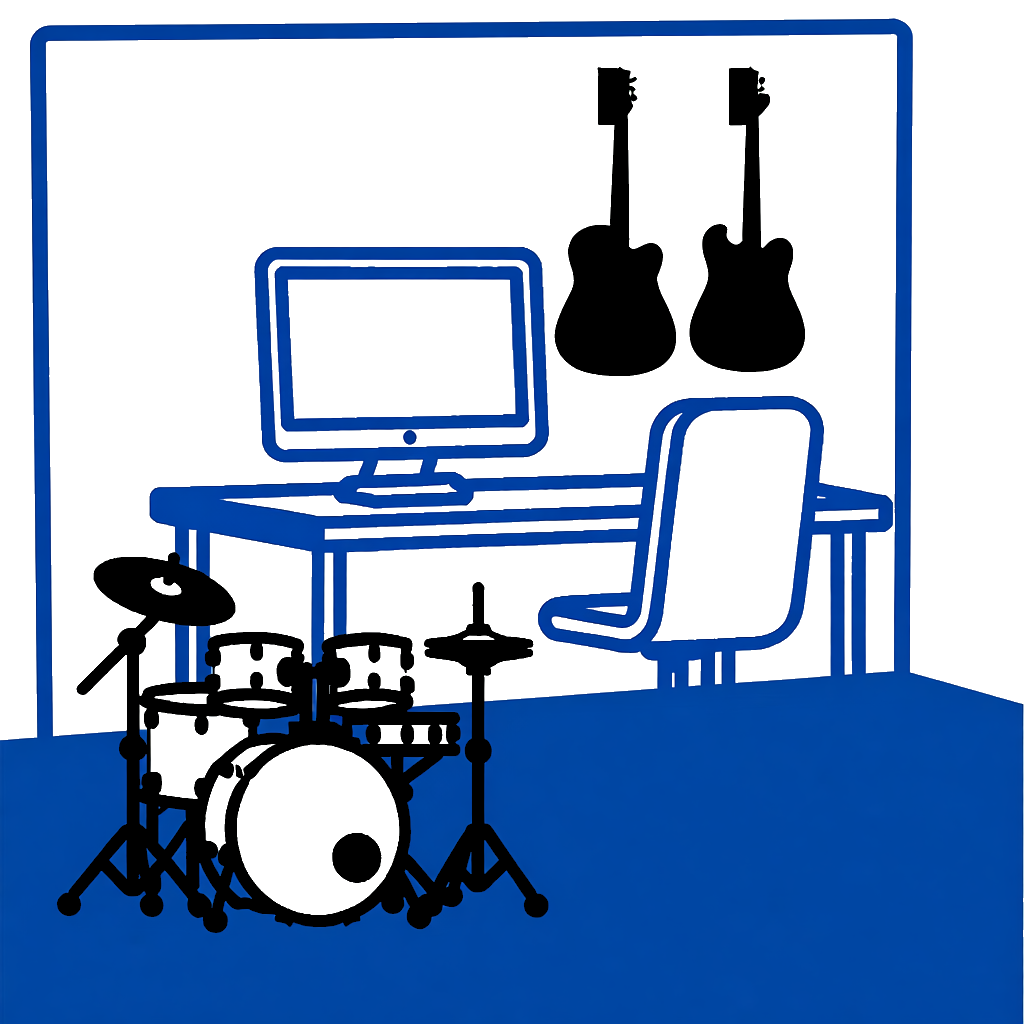}
            \caption{T29: Guitars added on wall.} 
        \end{subfigure}
    \end{minipage}
    }
    \caption{A representative example of incremental common-ground externalization. Left: verbatim dialogue turns together with Observer decisions and a cross-scene topological relation. Right: successive visual artifacts produced for the same scene. \legendsquarehex{1f4e79} Blue outlines mark assumptions, allowing extra inferred metadata such as room type to be encoded in the image.}
    \label{fig:trace_and_constructor_sequence}
\end{figure*}

In the visual condition, the Constructor produces schematic artifacts rather than photorealistic scenes. This choice is deliberate. Situated dialogue often contains underspecification, uncertainty, and incomplete evidence; a photorealistic rendering would force the model to hallucinate unsupported detail and thereby pollute the established common ground. We term this \textbf{\emph{Style-as-Semantics}}: a design principle in which visual style directly encodes epistemic status. Schematic rendering is used deliberately -- minimalist icons omit irrelevant detail that would otherwise force commitment to ungrounded content~\cite{larkin1987diagram}; color-coded outlines visually distinguish what is certain, uncertain, or assumed, grounded in the principle that perceptual encodings can directly signal uncertainty~\cite{maceachren2012semiotics, padilla2021uncertainty} (see Figure~\ref{fig:trace_and_constructor_sequence}). The color-coded outlines of the objects rendered denote the following:
\textbf{Black -} Confirmed entities with grounded positions; 
\textbf{\textcolor{red}{Red}~-}~Confirmed entities with unresolved position; 
\textbf{\textcolor{blue}{Blue} -} Assumptions not explicitly confirmed in the dialogue.

This design keeps uncertainty explicit rather than collapsing it into a single over-committed scene interpretation, while clearly separating what has been established from what remains assumed. More broadly, the artifact externalizes the current interpretation of the dialogue while preserving where that interpretation remains uncertain. It makes it easier for downstream reasoning to distinguish grounded facts from assumptions and uncertainties.

\textbf{The Linker (L)}: Individual artifacts capture local scene interpretations, but persistent grounding also requires tracking relations \emph{between} scenes, such as revisitation, movement, and topological transitions. We therefore introduce a \textit{Linker} that extracts sparse cross-artifact relations from the metadata $m_t$ and 
the dialogue context window $H_t$ surrounding the current frame. These relations take the form of triplets $\tau_t$, for example
$\langle \texttt{B\_3}, \text{is\_north\_of}, \texttt{B\_2} \rangle$ where \texttt{B\_2} and \texttt{B\_3} are artifact ids.

The Linker plays a complementary role to the Constructor. While the artifacts preserve local perceptual and relational structure within a scene, the triplets preserve abstract continuity across scenes.

\subsection{Inference over Grounded Artifacts}

At inference time, a \textbf{Reasoner (R)} receives a natural language 
question $Q$ and produces an answer $\hat{y}$ by retrieving from the externalized memory bank $\mathcal{M}$ via retrieval queries $q$. The goal of this phase is not simply to recover previously stored information, but to test whether explicit artifacts help the model resolve references, preserve distinctions, and reason over dialogue state more effectively than relying on the raw transcript alone.

\textbf{Planned RAG}: Given $Q$, the Reasoner first produces a plan $\mathcal{P} = \{p_1, \dots, p_L\}$,  where each step $p = (c, i)$ pairs a command $c$ with an instruction $i$. The commands are:
\begin{itemize}
    \item \texttt{POV}, to filter evidence according to speaker perspective,
    \item \texttt{RAG[N]}, to retrieve $N$ relevant artifacts; a retrieval query $q$~(here, $i==q$) is issued to $\mathcal{M}$ if and only if $c = \texttt{RAG}$,    \item \texttt{PROCESS}, to perform intermediate reasoning over the retrieved evidence, and
    \item \texttt{FINAL\_ANSWER}, to synthesize the final response.
\end{itemize}

Each operation is paired with an instruction specifying the required action, for example, ``RAG[5] bedrooms with white walls". This decomposition aligns the reasoning process more closely with the demands of grounded dialogue tasks. Queries in IndiRef often require more than simple lexical matching. The system may need to identify the relevant point in the dialogue history, select the correct speaker perspective, retrieve the appropriate local scene, and then reason over both local and cross-scene evidence. The plan $\mathcal{P}$ therefore acts as a controller for how the Reasoner interrogates externalized common ground, with retrieval being selectively triggered only when $c = \texttt{RAG}$.

The \textbf{Visual Condition} retrieval is performed with a hybrid scoring function that combines visual similarity of $q$ against the artifact itself with text similarity against its associated metadata:
\begin{equation*}
    \text{Score}(\mathcal{I}_k, q) =
    \lambda \cdot \cos(\Psi_v(\mathcal{I}_k), \Psi_t(q))
    +
    (1-\lambda) \cdot \cos(\Psi_t(m_k), \Psi_t(q))
    \label{eq:retrieval_viz_score}
\end{equation*}
where $\Psi_v$ and $\Psi_t$ are the vision and text encoders respectively, $m_k$ denotes the metadata associated with artifact $\mathcal{I}_k$, and $\lambda \in [0,1]$ balances depictive evidence against textual fallback. We set $\lambda = 0.7$, thereby prioritizing pixel-level evidence while retaining metadata as a complementary retrieval signal. The retrieved evidence is then passed to the Reasoner in three parts: the artifact itself, its metadata $m_k$, and the cross-scene triplets $\tau_k$ containing the retrieved artifacts. In the \textbf{Textual Condition}, retrieval is performed over the scene summaries $\mathcal{S}_k$ using text-to-text similarity with $\tau_k$.

Finally, the Reasoner synthesizes a predicted answer $\hat{y}$ from the retrieved evidence, and a \textbf{Judge (J)} evaluates whether $\hat{y}$ matches the gold answer $y^*$ as a binary outcome. For evaluation, we adopt the LLM-as-a-Judge protocol of \citet{mohapatra2026frame}, who show that this answer-matching task is straightforward for a competent LLM and achieves high agreement with human annotators.

\subsection{Implementation Details} \label{sec:implementation}

We instantiate the pipeline with Qwen-family foundation models. The \textit{Observer}, the verifier used by the \textit{Constructor}, the \textit{Linker}, and the \textit{Reasoner} are all implemented with \textbf{Qwen3-VL-32B-Thinking}~\citep{bai2025qwen3vl}. Using a common multimodal backbone across these modules reduces architectural variability and ensures that both artifact construction and downstream reasoning are carried out within the same instruction-following framework.

For visual artifact generation, we use an 8-step Lightning LoRA variant of \textbf{Qwen-Image-Edit}~\citep{wu2025qwenimage}\footnote{\href{https://github.com/ModelTC/Qwen-Image-Lightning/}{Qwen-Image-Edit-2509-Lightning-8steps-V1.0}}. This helps with iterative scene editing as the $\textsc{Continue}$ action must preserve previously grounded content while integrating new information. Further, we found that tiny pixel changes built up across repeated edits since each generation step slightly altered regions that should have stayed uniform. To prevent this, we added a normalization step after each generation that maps near-uniform pixels to canonical colors, which stabilized the sequence and preserved information over long dialogues.

For retrieval, we use a CLIP-style~\cite{clip}\footnote{\href{https://huggingface.co/sentence-transformers/clip-ViT-L-14}{https://huggingface.co/sentence-transformers/clip-ViT-L-14}} dual encoder to compute the similarity scores. During artifact construction, the rejection-sampling loop uses $g=3$ candidate generations. We used Llama 3.1 8B~\citep{llama3} for LLM as a Judge. 

All experiments were conducted on three NVIDIA A100 (80GB) GPUs. Qwen3-VL-32B-Thinking was deployed across two GPUs using tensor parallelism, while Qwen-Image-Edit was hosted on a separate GPU. All prompts available in Appendix \ref{apx:prompts}.

\section{Results} \label{sec:results}

We evaluate the proposed framework on the IndiRef benchmark ($N=50$ per semantic category; $N_{total}=200$). We compare two incremental agentic conditions, \textit{Agentic-Image} and \textit{Agentic-Text}, against two non-incremental baseline \textbf{\textit{Full Dialogue}~(FD)} settings which answer directly from the full transcript in the context using: (i) the same VL model used for our agentic framework, and (ii) similar text-only thinking model \textit{Qwen-QwQ-32B} \citep{mohapatra2026frame}. 

\begin{table}[htbp]
    \centering
    \resizebox{\columnwidth}{!}{%
    \begin{tabular}{lcccc}
        \toprule
        \multirow{2}{*}{Framework} & \multicolumn{4}{c}{Relation Type (Accuracy)} \\ 
        \cmidrule(lr){2-5}
        & Temporal & Spatial & Attributive & Inferred \\ \midrule
        FD (Qwen3-VL-Thinking) & 0.18 & 0.10 & 0.20 & 0.24 \\
        FD (Qwen-QwQ) & 0.32 & \textbf{0.38} & 0.40 & 0.40 \\
        \hdashline
        Agentic-Image & \textbf{0.50} & 0.24 & \textbf{0.44} & \textbf{0.58} \\
        Agentic-Text & 0.42 & 0.26 & \textbf{0.44} & 0.46 \\
        \bottomrule
    \end{tabular}
    }
    \caption{Accuracy across relation types.}
    \label{tab:results_comparison}
\end{table}

Table~\ref{tab:results_comparison} reveals several key findings. First, both agentic variants outperform the \textit{Full Dialog} Qwen3-VL baseline across all relation types, suggesting that \emph{incremental externalization} brings some benefits: structuring dialogue into discrete, retrievable frames reduces contextual drift and representational blur. Notably, this pattern contrasts with the findings of \citet{mohapatra2026frame}, where the \textit{Full Dialog} setting was stronger than the earlier agentic representation pipeline. This reversal suggests that the critical factor is not agentic decomposition by itself, but \emph{incrementality}: constructing the dialogue state progressively as the interaction unfolds appears more effective than generating a representation retrospectively from the complete transcript. Second, in the \textit{Full Dialog} setting, the text-only \textit{Qwen-QwQ-32B} model outperforms Qwen3-VL-Thinking-32B, underscoring the weaker reasoning ability of the VL model on raw long-context dialogue. Despite this, our \textit{Agentic-Image} framework, built primarily on the same VL backbone, exceeds \textit{Qwen-QwQ-32B} on \textit{Temporal}, \textit{Attributive}, and \textit{Inferred} queries. A closer analysis suggests that many \textit{Spatial} questions in IndiRef require cross-frame reasoning. The weaker performance of \textit{Agentic-Image} on this category therefore points less to a limitation of local visual grounding and more to a bottleneck in the \textit{Linker}, whose cross-frame topological relations were often insufficient to support reliable traversal.

Further, the comparison between \textit{Agentic-Image} and \textit{Agentic-Text} reveals a structured modality trade-off rather than a uniform advantage for either condition. \textit{Agentic-Image} performs best on \textit{Temporal} (0.50) and \textit{Inferred} (0.58) relations, and ties for the top score on \textit{Attributive} queries (0.44). By contrast, \textit{Agentic-Text} performs slightly better on \textit{Spatial} relations (0.28 vs.\ 0.24). This pattern is consistent with our hypothesis: visual artifacts are particularly helpful when grounding depends on preserving concrete scene commitments over time, retaining fine-grained perceptual distinctions, and supporting multi-step inference over evolving state. Textual externalization, however, remains competitive when the queried information is more naturally preserved as explicit propositions, particularly for cross-room topology and movement relations.


\begin{figure}[!h]
    \centering
    \includegraphics[width=1\linewidth]{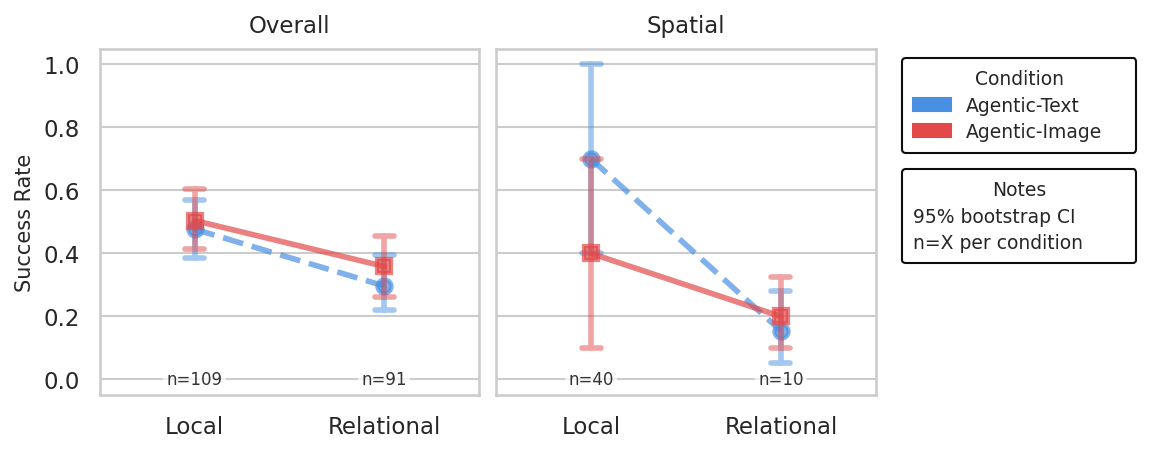}
    \caption{Performance by reasoning scope. Left: overall accuracy. Right: accuracy within the Spatial question type.
    }
    \label{fig:reasoning_scope}
\end{figure}

\begin{figure*}[!hbtp]
    \centering
    \resizebox{0.7\linewidth}{!}{%
    \begin{minipage}[t]{0.5\textwidth}
        \vspace{0pt} 
        \begin{dialoguebox}[width=0.95\columnwidth]{Verbatim Trace Snippet (ID: 14\_116\_47\_95)}
            \begin{dialoguelist}
                \item \textbf{[Turn 15] A}: I am in the basement \hfill ($\textsc{New} \rightarrow \texttt{A\_5}$)
                \item \textbf{[Turn 16] B}: I'm in a basement. \hfill ($\textsc{New} \rightarrow \texttt{B\_4}$)
                \item \textbf{[Turn 17] B}: Has a white staircase? \hfill ($\textsc{Continue} \rightarrow \texttt{B\_4}$)
                \item \textbf{[Turn 18] A}: no \hfill ($\textsc{Skip}$)
                \item \textbf{[Turn 19] A}: mine has wooden brown stair case \hfill ($\textsc{Continue} \rightarrow \texttt{A\_5}$)
                \item \textcolor{violet}{\textbf{B}: What was the color of the stair in my basement like?}
                \item \textcolor{purple}{\textbf{A}: It was white}
                \item \textbf{Image Condition Answer}: white \hfill ($\textsc{Correct}$)
                \item \textbf{Text Condition Answer}: brown \hfill ($\textsc{Incorrect}$)
            \end{dialoguelist}
        \end{dialoguebox}
        \label{fig:qual_implicit_trace}
    \end{minipage}\hspace{0.05\columnwidth}%
    \begin{minipage}[t]{0.45\textwidth}
        \vspace{0pt} 
        \begin{subfigure}[t]{0.49\linewidth}
            \centering
            \includegraphics[width=\linewidth]{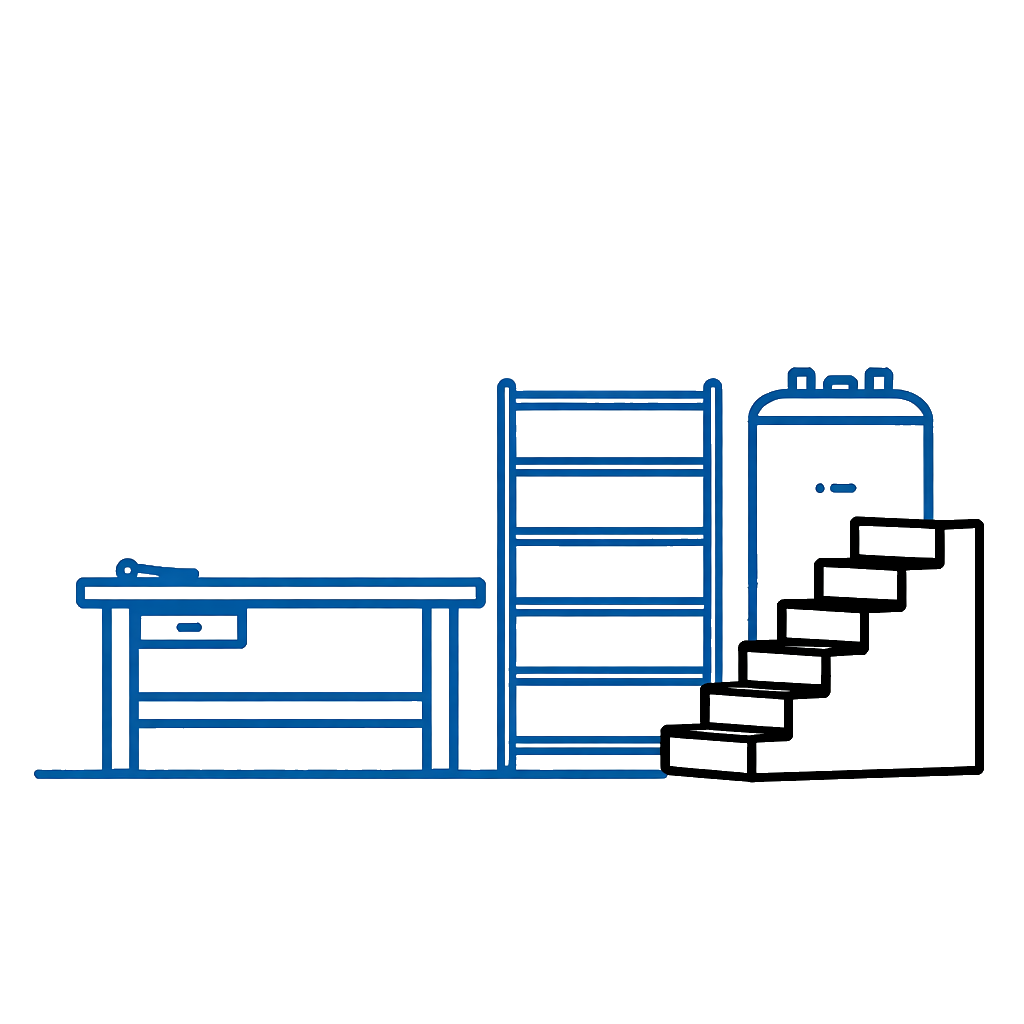}
            \caption{Turn 17. A's current POV of B's environment}
            \label{fig:commitment_17}
        \end{subfigure}\hfill
        \begin{subfigure}[t]{0.49\linewidth}
            \centering
            \includegraphics[width=\linewidth]{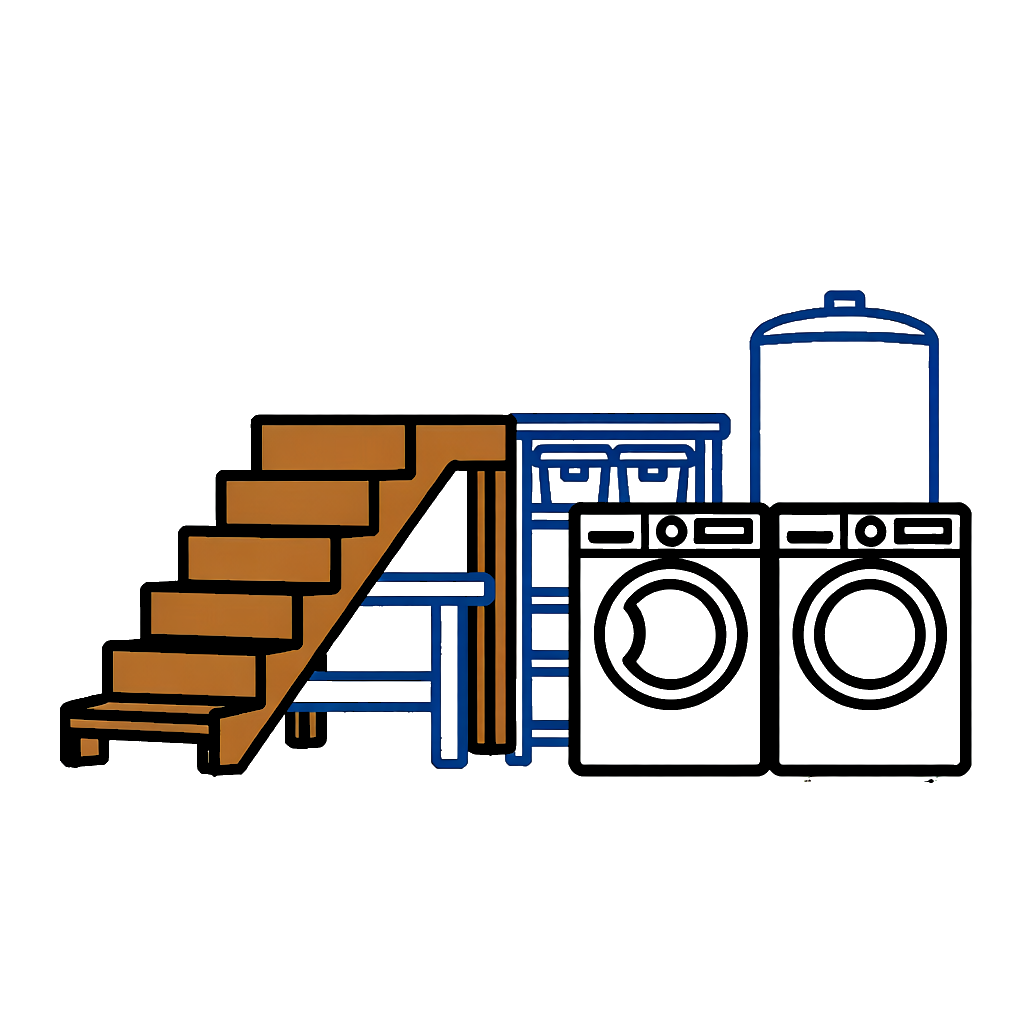}
            \caption{Turn 19. A's POV of A's own env. that it shared with B.}
            \label{fig:commitment_19}
        \end{subfigure}
        
        \label{fig:commitment}
    \end{minipage}
    }
    \caption{Left: B uses a clarification question to check if A shares the same environment, implicitly revealing what B sees. Right: Visual scaffolding enforces epistemic boundaries. When B asks a clarification question, a white staircase is generated exclusively in B's environment (a). A's subsequent correction updates only A's distinct canvas (b), preventing cross-contamination.}
    \label{fig:stairs}
\end{figure*}

\section{Analysis} \label{sec:analysis}

The quantitative divergence between the visual and textual conditions highlights specific modality-driven trade-offs in multimodal memory management. We decompose these results to diagnose the underlying mechanisms of success and failure.

\paragraph{\textbf{Performance by Question Typology.}}
A first useful distinction is between queries that require reasoning \textit{within} a single retrieved scene (\textit{Local}) and queries that require reasoning \textit{across} scenes (\textit{Relational} -- 
including spatial distance, movement, revisitation). When performance is grouped by reasoning scope, a clear pattern emerges: \textit{Local} questions are consistently easier than \textit{Relational} ones (Figure~\ref{fig:reasoning_scope}, left).\footnote{The distiction in reasoning scope has been manually annotated by the author and vetted via GPT-4o-mini~\cite{openai2024gpt4ocard}. Prompt available in Appendix \ref{apx:prompts}, Figure \ref{prompt:annotator}.} Across conditions, local questions reach roughly $50\%$ accuracy, whereas relational questions fall closer to $30$--$35\%$.

This distinction is especially revealing in the spatial domain (Figure~\ref{fig:reasoning_scope}, right) which contains numerous inter-scene questions (eg. \textit{``room to the left of kitchen''}). Both conditions degrade when the task shifts from identifying properties of a room to first traversing the environment and consequently gather information, but the drop is sharper in the image condition. The textual condition retains a modest advantage because scene summaries often preserve transitions propositionally, e.g., by explicitly stating that a speaker \textit{left the bathroom and went north}. In contrast, the visual pipeline represents rooms as discrete artifacts and must recover connectivity through the sparse triplets produced by the \textit{Linker}. When those links are missing, weak, or not retrieved strongly enough, the system fails to point out the correct artifact. This gap suggests that navigating over a fragmented memory bank and preserving their relations over time is a more challenging task especially for visual artifacts.


\begin{figure}[!htbp]
    \centering
    \includegraphics[width=1\linewidth]{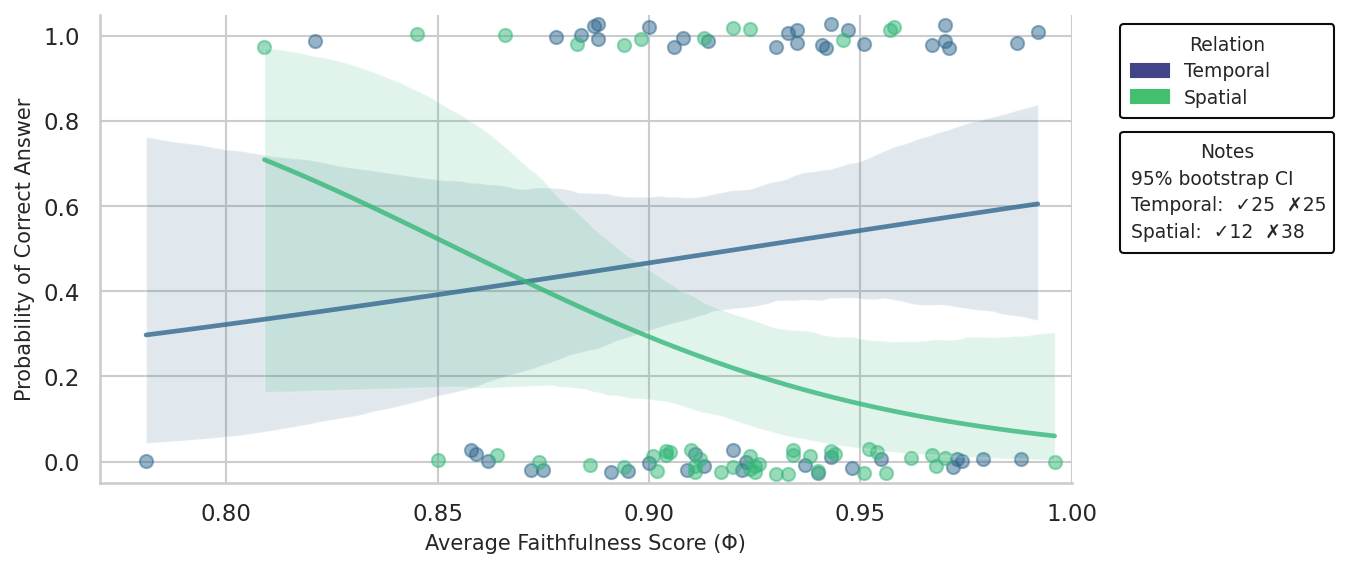}
    \caption{Logistic Regression showing probability of a correct answer as a function of Visual Faithfulness ($\Phi$).
    \label{fig:faithfulness_corr}
}
\end{figure}

This interpretation is further supported by the relationship between the Constructor's visual faithfulness score $\Phi$ and downstream accuracy (Figure~\ref{fig:faithfulness_corr}). For \textit{Temporal} queries, the relation is positive: as frame faithfulness increases, answer accuracy also improves. This is consistent with the intended role of visual scaffolding. When the system builds a faithful frame sequence, temporal information is preserved directly in the sequence itself, forming a robust timeline that is easy to retrieve from without a separate \textit{Linker}.

For \textit{Spatial} queries, however, the trend reverses: higher local faithfulness is associated with \emph{lower} accuracy. The reason is a retrieval mismatch. A highly faithful image of a room is a strong retrieval target: when a query mentions a salient object in that room, the multimodal retriever confidently returns that image. But many spatial questions ask not about that room itself, but about an adjacent room or a topological relation 
between rooms. The system locks onto the most visually similar artifact -- the correct starting point -- but never follows the cross-frame link needed to reach the actual answer.

\paragraph{\textbf{Mechanisms Behind the Gains of Visual Scaffolding.}}
The advantages of the image condition on \textit{Temporal}, \textit{Attributive}, and especially \textit{Inferred} questions can be traced to two related mechanisms: reduced representational blur and forced commitment to a concrete scene interpretation.

\begin{figure}[b]
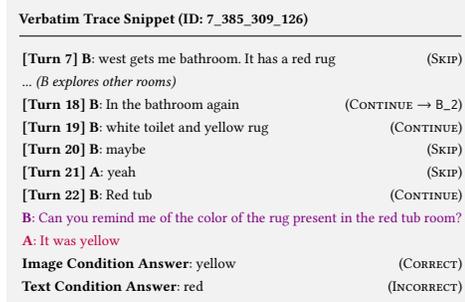

    \centering
    \resizebox{0.7\columnwidth}{!}{%
    \begin{minipage}{1\columnwidth}
        \begin{dialoguebox}{Verbatim Trace Snippet (ID: 7\_385\_309\_126)}
            \begin{dialoguelist}
                \item \textbf{[Turn 7] B}: west gets me bathroom. It has a red rug \hfill ($\textsc{Skip}$)
                \item ... \textit{(B explores other rooms)}
                \item \textbf{[Turn 18] B}: In the bathroom again \hfill ($\textsc{Continue} \rightarrow \texttt{B\_2}$)
                \item \textbf{[Turn 19] B}: white toilet and yellow rug \hfill ($\textsc{Continue}$)
                \item \textbf{[Turn 20] B}: maybe \hfill ($\textsc{Skip}$)
                \item \textbf{[Turn 21] A}: yeah \hfill ($\textsc{Skip}$)
                \item \textbf{[Turn 22] B}: Red tub \hfill ($\textsc{Continue}$)
                \item \textcolor{violet}{\textbf{B}: Can you remind me of the color of the rug present in the red tub room?}
                \item \textcolor{purple}{\textbf{A}: It was yellow}
                \item \textbf{Image Condition Answer}: yellow \hfill ($\textsc{Correct}$)
                \item \textbf{Text Condition Answer}: red \hfill ($\textsc{Incorrect}$)
            \end{dialoguelist}
        \end{dialoguebox}
    \end{minipage}
    }%
    \caption{B's exploration of the bathroom involves conflicting color attributes across distant turns. 
    }
    \label{fig:qual_blur_trace}
\end{figure}

A recurring failure mode of text-based externalization is \emph{representational blur}, where distinct entities or attributes are compressed into a vague or partially conflated summary. This is especially problematic in IndiRef, where many questions hinge on small perceptual distinctions across revisited or highly similar scenes. 
Consider Figure \ref{fig:qual_blur_trace}: the two bathrooms encountered by B (one with a red rug, the other with a yellow rug and a red tub, mentioned across distant turns) create the conditions under which textual compression fails, as the summary merges these cues into a collapsed description and produces a wrong answer. The visual scaffold avoids this collapse by instantiating both attributes as distinct elements within a single explicit state -- not merely adding detail, but making collapse structurally impossible.

A second advantage of visual scaffolding is that artifact construction imposes a form of \emph{forced commitment}. When the two participants occupy separate environments, a local correction by one speaker can cause text-based summarizers to incorrectly overwrite the other speaker's established state.
We observe this dynamic in Figure \ref{fig:stairs}, where Player B implicitly reveals their local state by asking a clarification question. During evaluation, the Reasoner is given A's role and is asked: \textit{``What was the color of the stair in my basement like?''} by B. The textual summarizer frequently interprets Player A's denial as a global correction of a mistaken fact, explicitly overwriting Player B's reality with Player A's description. Consequently, the text Reasoner incorrectly answers: \texttt{``brown''}.

The visual scaffold bypasses this vulnerability. When B asks \textit{``Has a white staircase?''} (Turn 17), the framework routes this update exclusively to B's active visual frame, compelling the Constructor to render the white staircase. A's subsequent assertion updates only A's distinct canvas (Figure \ref{fig:stairs}, right). Because the visual architecture enforces a strict physical separation of the two agents' environments, it prevents linguistic cross-contamination and correctly retrieves B's original attribute (\texttt{``white''}). This also suggests why Agentic-Image performs better in Inferred cases.

\paragraph{\textbf{Pragmatic Limitations and Failure Modes}}
Despite these benefits, the visual framework also exposes important limitations. The first concerns \emph{non-depictable} or pragmatically indirect information. The system is effective at externalizing positive, visually expressible commitments, but it is substantially weaker when the relevant information is conveyed through negation, denial, uncertainty, or implicature. Figure~\ref{fig:qual_negative_trace} illustrates this problem. Speaker B's question, \textit{``It's not yellow?''}, pragmatically reveals information about B's own room while checking whether it matches A's. However, because the update is framed negatively and indirectly, the Observer treats it as non-depictable and skips it. The result is a lost grounding signal: the information never enters the artifact bank, and the later query fails. This is a broader limitation of the current framework. Images are effective carriers of positive depictable commitments, but they do not natively encode absence, negation, or discourse-level qualifications unless these are explicitly preserved in metadata and made retrievable downstream.

\begin{figure}[!tbp]
    \centering
    \resizebox{0.7\columnwidth}{!}{%
    \begin{minipage}{1\columnwidth}
        \begin{dialoguebox}{Verbatim Trace Snippet (ID: 360\_98\_113\_352)}
            \begin{dialoguelist}
                \item \textbf{[Turn 0] A}: Hi there. Looks like I'm outside \hfill ($\textsc{New} \rightarrow \texttt{A\_1}$)
                \item \textbf{[Turn 1] B}: im in a bathroom \hfill ($\textsc{New} \rightarrow \texttt{B\_1}$)
                \item \textbf{[Turn 2] B}: im in a childs room with a pink moon \hfill ($\textsc{New} \rightarrow \texttt{B\_2}$)
                \item \textbf{[Turn 3] A}: I made it to the childs room. Mine has a blue wall \hfill ($\textsc{New} \rightarrow \texttt{A\_2}$)
                \item \textbf{[Turn 4] A}: OK, can you find this room with blue wall? \hfill ($\textsc{Skip}$)
                \item \textbf{[Turn 5] B}: It's not yellow? \hfill ($\textsc{Skip}$)
                \item \textbf{[Turn 6] B}: ill find you \hfill ($\textsc{Skip}$)
                \item \textbf{[Turn 7] A}: ok \hfill ($\textsc{Skip}$)
                \item \textcolor{violet}{\textbf{B}: What was the colour of the walls of the pink moon room?}
                \item \textcolor{purple}{\textbf{A}: It was yellow}
                \item \textbf{Image Condition Answer}: white \hfill ($\textsc{Incorrect}$)
                \item \textbf{Text Condition Answer}: not specified \hfill ($\textsc{Incorrect}$)
            \end{dialoguelist}
        \end{dialoguebox}
    \end{minipage}
    }%
    \caption{B uses a negative clarification question (\textit{``Not yellow?''}) to implicitly state the color of their own room. The Observer treats this as non-depictable ($\textsc{Skip}$).}
    \label{fig:qual_negative_trace}
\end{figure}

The second limitation occurs during memory retrieval. Human communication relies heavily on indirect requests; for instance, \textit{``Do you remember [X]?''} functions as a prompt to state \textit{[X]}. For example, when asked, \textit{``Do you remember the color of the walls of my first room?''}, execution logs reveal that the \textit{Agentic-Image} Reasoner successfully retrieved the correct visual artifact alongside its exact metadata (\texttt{``red room''}). Yet, the final output was simply: \texttt{``yes''}. 


\section{Hybrid Setting}
The qualitative analysis suggests that the contrast between \textit{Agentic-Image} and \textit{Agentic-Text} is best understood not as a simple competition between two memory formats, but as a division of labor. The image condition is stronger when grounding depends on preserving fine-grained perceptual distinctions, maintaining concrete scene commitments over time, and supporting inference from explicit visualized states. The text condition, by contrast, remains more reliable for topology, negation, uncertainty, and other abstract or non-depictable aspects of the shared state. These complementary strengths motivated a final experiment where rather than forcing the system to rely on a single externalization format, we introduced an \textit{Agentic-Both} condition that maintains both a visual artifact and its paired textual representation for each frame, and retrieves from both at inference time and sends them together as context to our VL model.

\begin{table}[htbp]
    \centering
    \begin{tabular}{lcccc}
        \toprule
        \multirow{2}{*}{Framework} & \multicolumn{4}{c}{Relation Type (Accuracy)} \\ 
        \cmidrule(lr){2-5}
        & Temporal & Spatial & Attributive & Inferred \\ \midrule
        Agentic-Image & 0.50 & 0.24 & 0.44 & 0.58 \\
        Agentic-Text & 0.42 & 0.26 & 0.44 & 0.46 \\
        \hdashline
        Agentic-Both & 0.58 & 0.44 & 0.64 & 0.52 \\
        \bottomrule
    \end{tabular}
    \caption{Accuracy of Agentic-Both setting.}
    \label{tab:results_agentic_both}
    \vspace{-5mm}
\end{table}

The results from Table~\ref{tab:results_agentic_both} show that \textit{Agentic-Both} produced the best overall performance across most categories. These gains suggest that the two modalities preserve partially non-overlapping evidence about the evolving common ground. In particular, the hybrid setting appears to reduce two of the main failure modes identified earlier: representational blur in the text condition and the navigation bottleneck in the image condition. The only exception is \textit{Inferred} grounding, where \textit{Agentic-Image} remains slightly stronger than the hybrid setting (0.58 vs.\ 0.52). One possible explanation is that many inferred questions benefit precisely from the \emph{forced commitment} induced by visual externalization. Adding the textual channel may reintroduce ambiguity, thereby slightly diluting the benefit of the image-based commitment.

\section{Conclusion and Future Work}
\label{sec:conclusion}

This paper showed that conversational grounding in conversational agents benefits from \emph{incremental externalization}: representing dialogue as discrete, content-addressable states is more effective than reasoning directly over the full transcript. Within this incremental setting, visual scaffolding provides crucial grounding ability by forcing concrete scene commitments, reducing representational blur, and better preserving fine-grained perceptual distinctions over time. At the same time, the results show that visual scaffolding alone is not a complete substitute for textual representations. The image condition remains weaker where success often depends on explicit cross-frame topology rather than local depictive accuracy. More broadly, the two modalities exhibit complementary strengths: images are effective for preserving concrete perceptual commitments, while text remains better suited to abstract, relational, and non-depictable information. This complementarity is confirmed by the \textit{Agentic-Both} setting, which achieves the strongest overall performance and suggests that persistent common ground is best modeled as an explicitly multimodal structure.

These findings motivate three main directions for future work. First, rather than combining text and images only at retrieval time, future systems should build a more integrated multimodal memory that jointly represents visual artifacts, textual summaries, and relational links. Second, low-confidence or conflicting multimodal evidence should be connected more directly to dialogue policy so that uncertainty can trigger clarification and repair rather than silent commitment. Our visual scaffolding method provides a way for the model to generate a mental imagery to detect such conflicting or underspecified information. Third, the main bottleneck for spatial grounding while using visual scaffolding lies in \emph{link quality}, which motivates stronger relational representations and navigation-aware retrieval, potentially through structured alternatives.

\section{Safe and Responsible Innovation Statement}

This work aims to improve the reliability of multimodal dialogue systems by making their intermediate state explicit and inspectable. A key potential benefit is safer and more natural human--machine interaction, since stronger grounding can reduce ambiguity and unreliable behavior. However, such systems still raise concerns around privacy, bias, and overconfident interpretation of user intent. Our experiments use benchmark data rather than personal user data, and we view this work as a step toward safer multimodal interaction through more transparent state tracking. Future deployment should include privacy-conscious practices, bias evaluation, and mechanisms for uncertainty-aware clarification.

\bibliographystyle{ACM-Reference-Format}
\bibliography{sample-base, references}

\newpage
\appendix

\section{Example of Pipeline} \label{apx:example_pipeline}

To illustrate the transformation from dialogue to visual artifacts, we examine a trace from the test set involving a description of a bedroom in Figures~\ref{fig:turn1}, \ref{fig:turn2}, \ref{fig:turn4}, and \ref{fig:turn5}. This trace demonstrates the \textit{Observer}'s state segmentation and the \textit{Constructor}'s iterative refinement capabilities.

\begin{figure}[!htbp]
    \centering
    
    \begin{dialogueboxE}[0.9\linewidth]{Turn 1: Initialization [NEW]}
        \footnotesize
        \begin{dialoguelist}
            \item \textbf{Utterance (B):} ``in bedroom''
            \item \textbf{Action:} The \textit{Observer} detects a location change and triggers a \textsc{New} action (Frame ID: $B\_1$).
            \item \textbf{Constructor:} Since there is no prior visual context, the Constructor generates a base scene from scratch.
            \item \textit{Prompt:} ``A clean, minimalist, iconic scene. In a bedroom, a bed (in blue outline), a nightstand (in blue outline), and a lamp (in blue outline). Solid white background, no shadows.''
            \item \textit{Result:} A foundational image establishing the spatial layout.
        \end{dialoguelist}
    \end{dialogueboxE}
    \caption{Turn 1: Initialization of the visual state based on a location change. The Observer detects the need for a new frame, and the Constructor generates an initial scene with the specified objects.}
    \label{fig:turn1}
\end{figure}

\vspace{0.5cm}

\begin{figure}[!htbp]
    \centering

    \begin{dialogueboxE}[0.9\linewidth]{Turn 2: Incremental Update [CONTINUE]}
        \footnotesize
        \begin{dialoguelist}
            \item \textbf{Utterance (B):} ``window at back''
            \item \textbf{Action:} The \textit{Observer} recognizes this as an elaboration of the current state ($B\_1$) and triggers \textsc{Continue}.
            \item \textbf{Constructor:} The system enters ``Editing Mode.'' It retrieves the image from Turn 1 and applies a structural update.
            \item \textit{Prompt:} ``Add a wall (in blue outline) at the back of the bedroom... Add a window (in black outline) on the wall. \textbf{Keep} the bed, nightstand, and lamp unchanged.''
            \item \textit{Result:} The visual state ($B\_1\_seq2$) now contains the window while preserving the geometric consistency of the previous objects.
        \end{dialoguelist}
    \end{dialogueboxE}
    \caption{Turn 2: The Observer detects an elaboration of the current scene and triggers a continuation. The Constructor modifies the existing image by adding new elements while keeping the previous ones intact, demonstrating the system's ability to maintain visual consistency across updates.}
    \label{fig:turn2}
\end{figure}

\vspace{0.5cm}

\begin{figure}[!t]
    \centering
    \begin{dialogueboxE}[0.9\linewidth]{Turn 4: Phatic Filtering [SKIP]}
        \footnotesize
        \begin{dialoguelist}
            \item \textbf{Utterance (A):} ``Okay, I'm still looking.''
            \item \textbf{Action:} The \textit{Observer} classifies this as non-depictable phatic communication.
            \item \textit{Result:} No visual generation occurs. The system maintains the state at $B\_1\_seq3$ (generated in Turn 3), preventing the visual history from being cluttered with redundant frames.
        \end{dialoguelist}
    \end{dialogueboxE}
    \caption{Turn 4: The Observer identhifies a phatic utterance that does not contribute new visual information. By classifying it as \textsc{Skip}, the system avoids unnecessary generation, demonstrating its ability to filter out non-visual content and maintain a clean visual history.}
    \label{fig:turn4}
\end{figure}

\vspace{0.5cm}

\begin{figure}[!t]
    \centering
    \begin{dialogueboxE}[\linewidth]{Turn 5: Feature Modification [CONTINUE]}
        \footnotesize
        \begin{dialoguelist}
            \item \textbf{Utterance (B):} ``red and grey stripped bedspread''
            \item \textbf{Action:} The user refines a specific attribute of an existing object.
            \item \textbf{Constructor:} The prompt explicitly targets the specific entity for replacement while freezing the rest of the scene.
            \item \textit{Prompt:} ``\textbf{Replace} the bed (in blue outline) with a bed (in black outline) that has a red and grey striped bedspread. \textbf{Keep} the nightstand, lamp, wall... unchanged.''
            \item \textit{Result:} The artifact $B\_1\_seq4$ reflects the texture update. By explicitly instructing the model to ``Keep'' other elements, we mitigate the catastrophic forgetting often observed in iterative generation.
        \end{dialoguelist}
    \end{dialogueboxE}
    \caption{Turn 5: The user provides a detailed refinement of an existing object. The Constructor processes this as a targeted modification, replacing the bed's appearance while preserving the rest of the scene. This demonstrates the system's ability to handle incremental updates without losing previously established visual information.}
    \label{fig:turn5}
\end{figure}

\pagebreak
\
\newpage

\section{Prompts} \label{apx:prompts}

We further provide all the prompts used for the multimodal representation work, specifically Figures \ref{prompt:observer}, \ref{prompt:constructor}, \ref{prompt:summarizer}, \ref{prompt:verifier}, \ref{prompt:linker}, \ref{prompt:reasoner}, \ref{prompt:annotator}, and \ref{prompt:judge}.

\begin{figure}[h]
    \begin{observer}{Observer}
    \footnotesize
    You are an analytical and decisive AI, part of a common ground visualization mechanism. Your task is to analyze the conversation history and the current target utterance to decide how to interpret the information and update the mental imagery of the scene. \\

    \textbf{[Conversation Contextualization]} \\

    \textbf{Inputs:}
    \begin{itemize}
        \item \textbf{Context:} The conversation history containing previous utterances.
        \item \textbf{Previous Prompts:} A list of prompts that built the current mental imagery state.
        \item \textbf{Target Utterance:} The new piece of information to reason on.
    \end{itemize}

    \textbf{Output Template:}
\begin{verbatim}
{
  "frame_meta": "Scene Label: Non-visual info",
  "relation": "[RELATION_META]",
  "imagery": "[IMAGERY]",
  "action": "[ACTION]"
}
\end{verbatim}

    \textbf{Rules:}
    \begin{itemize}
        \item (Meta Information) \texttt{frame\_meta}: Identify the location gist (1-3 words) and append non-visual info (e.g., smells, temp). \texttt{relation}: Populate if the utterance indicates movement or connection to a previous frame.
        \item (Imagery) Extract only visualizable information from the \texttt{target}. Maintain the speaker's tone and use Theory of Mind to contextualize what the speaker sees.
        \item (Action) Decide the state transition:
        \begin{itemize}
            \item \texttt{[NEW]}: If \texttt{previous} is None OR visual info is unrelated to the current frame.
            \item \texttt{[CONTINUE]}: If visual info can be added to the current frame.
            \item \texttt{[SKIP]}: If no visual info is present (acknowledgments, fillers).
        \end{itemize}
    \end{itemize}

    \textbf{[Examples]}
\end{observer}
\caption{Condensed prompt for the Observer module. The full prompt includes detailed definitions for frame metadata and extensive conversation examples to guide the model's reasoning process regarding scene transitions and theory of mind.}
\label{prompt:observer}
\end{figure}


\begin{figure}[t]
\begin{drawer}{Constructor - Image Condition}
    \footnotesize
    You are a visual generation engine acting as either a "Scene Creator" (for new scenes) or a "Scene Editor" (for updates). Your goal is to translate text utterances into precise, renderable image prompts using a minimalist, iconic style. \\

    \textbf{[Conversation Contextualization]} \\

    \textbf{Inputs:}
    \begin{itemize}
        \item \textbf{Creation Mode:} \texttt{<context>}, \texttt{<target>}.
        \item \textbf{Editing Mode:} \texttt{<context>}, \texttt{<target>}, \texttt{<previous>} (history of prompts).
    \end{itemize}

    \textbf{Output Template:}
\begin{verbatim}
{
  "scene": "A clean, minimalist, iconic scene... [PROMPT Content]"
}
\end{verbatim}

    \textbf{Common Rules:}
    \begin{itemize}
        \item (Style) Clean, minimalist, iconic. Global background must be \textbf{Solid White}.
        \item (Structural Pairing) Background elements (e.g., walls) must include their logical pair (e.g., floors) in blue outline to ground the scene.
    \end{itemize}

    \textbf{Mode-Specific Rules:}
    \begin{itemize}
        \item \textbf{Creation (Start):}
        \begin{itemize}
            \item (Color Semantics) \textbf{Black Outline}: Explicit/Unambiguous objects. \textbf{Blue Outline}: Assumed/Context objects (limit 3). \textbf{Red Outline}: Uncertain/Relative position.
            \item (Composition) Total explicit + assumed objects should be between 1 and 3.
        \end{itemize}
        \item \textbf{Editing (Update):}
        \begin{itemize}
            \item (Operations) Use "Add", "Remove", "Replace". \textbf{Move} and \textbf{Rescale} must be deconstructed into \texttt{DELETE \$\$\$ ADD}.
            \item (Persistence) You must explicitly state which existing objects to "Keep unchanged".
            \item (Logic) If adding "A near B" and B is also new, add both instructions.
        \end{itemize}
    \end{itemize}

    \textbf{[Examples]}
\end{drawer}
\caption{Condensed prompt for the Constructor module in the Image condition. The full prompt includes comprehensive style guidelines for minimalist generation and complex editing logic—such as deconstructing move and rescale operations—to guide the model's reasoning process.}
\label{prompt:constructor}
\end{figure}


\begin{figure}[t]
\begin{drawer}{Constructor - Text Condition}
    \footnotesize
    You are a "Situation Summarizer" and "Tracker". Your goal is to create and maintain a concise, natural language "mental model" of the physical environment based on conversational updates. \\

    \textbf{[Conversation Contextualization]} \\

    \textbf{Inputs:}
    \begin{itemize}
        \item \textbf{Creation Mode:} \texttt{<target>}, \texttt{<context>} (to resolve ambiguities like "here" or "it").
        \item \textbf{Update Mode:} \texttt{<previous>} (current summary), \texttt{<target>} (new info).
    \end{itemize}

    \textbf{Output Template:}
\begin{verbatim}
{
  "scene": "A single, cohesive natural language 
            paragraph summarizing the state."
}
\end{verbatim}

    \textbf{Common Rules:}
    \begin{itemize}
        \item (Certainty Levels)
        \begin{itemize}
            \item \textbf{Explicit:} Clearly stated facts (e.g., "There is a red cup").
            \item \textbf{Implied:} Contextual deductions (e.g., "Kitchen implies stove").
            \item \textbf{Uncertain:} Vague info must be stated as such (e.g., "User believes there might be...").
        \end{itemize}
    \end{itemize}

    \textbf{Mode-Specific Rules:}
    \begin{itemize}
        \item \textbf{Creation:} Describe location, objects, and spatial relations based \textit{only} on the target, grounded by context.
        \item \textbf{Update:} Rewrite the previous summary to incorporate the target:
        \begin{itemize}
            \item \textbf{Add/Refine:} Integrate new objects or details.
            \item \textbf{Correct:} Overwrite contradictions (e.g., "Actually, it's blue").
            \item \textbf{Preserve:} Keep all non-contradicted previous facts.
        \end{itemize}
    \end{itemize}

    \textbf{[Examples]}
\end{drawer}
\caption{Condensed prompt for the Constructor module in the Text condition. The full prompt includes nuanced rules for handling uncertainty levels (explicit vs. implied) and examples of textual state updates to guide the model's reasoning process.}
\label{prompt:summarizer}
\end{figure}


\begin{figure}
\begin{various}{Verifier}
    \footnotesize
    \textbf{1. Prompt Summarizer (Ground Truth Extraction)}
    \begin{itemize}
        \item \textbf{Role:} Extract a comprehensive list of visual atomic facts from a sequence of prompts.
        \item \textbf{Rules:}
        \begin{itemize}
            \item (Filtering) Include objects with \textbf{Red} and \textbf{Black} outlines. Strictly \textbf{exclude Blue} outlines.
            \item (Atomicity) Break complex descriptions into single facts (e.g., "Red cat on sofa" $\rightarrow$ "There is a cat", "It is red", "On a sofa").
        \end{itemize}
        \item \textbf{Output:} JSON with a list of strings under key \texttt{"facts"}.
    \end{itemize}

    \vspace{0.3cm} \hrule \vspace{0.3cm}

    \textbf{2. Image Captioner (Visual Analysis)}
    \begin{itemize}
        \item \textbf{Role:} Describe an image following strict hierarchical rules based on outline colors.
        \item \textbf{Rules:}
        \begin{itemize}
            \item (Order) 1. Black Outlines $\rightarrow$ 2. Red Outlines $\rightarrow$ 3. Blue Outlines.
            \item (Detail Level) \textbf{Black/Red:} Comprehensive (Type, Color, Features, Location). \textbf{Blue:} Simple (Type only).
            \item (Relations) For Black objects, describe position relative to other Black objects.
        \end{itemize}
    \end{itemize}

    \vspace{0.3cm} \hrule \vspace{0.3cm}

    \textbf{3. Fact Checker (Visual Forensics)}
    \begin{itemize}
        \item \textbf{Role:} Verify if a list of text facts is true in the provided image.
        \item \textbf{Rules:}
        \begin{itemize}
            \item (Evidence First) For every fact, first attempt to draw a bounding box \texttt{[ymin, xmin, ymax, xmax]}.
            \item (Precision) Strict attribute matching (e.g., if fact says "red" but image is "blue", verdict is FALSE).
        \end{itemize}
        \item \textbf{Output:} JSON with \texttt{"fact"}, \texttt{"box"}, and \texttt{"verdict"} (boolean).
    \end{itemize}

    \textbf{[Examples]}
\end{various}
\caption{Condensed prompt for the Verifier module. The full prompt includes distinct instructions for the three sub-agents (Prompt Summarizer, Image Captioner, and Fact Checker) along with strict visual forensic rules to guide the model's reasoning process.}
\label{prompt:verifier}
\end{figure}


\begin{figure}[t]
\begin{various}{Linker}
    \footnotesize
    You are an analytical AI tasked with resolving natural language directives into precise \textbf{Knowledge Graph Triplets} that connect specific Frame IDs in a common ground visualization. \\

    \textbf{[Conversation Contextualization]} \\

    \textbf{Inputs:}
    \begin{itemize}
        \item \textbf{Directive:} The specific utterance describing the relation.
        \item \textbf{Context:} Frame IDs for \texttt{PREV\_FRAME}, \texttt{CURR\_FRAME}, and \texttt{NEXT\_FRAME}.
        \item \textbf{Frame Meta:} Scene information corresponding to these IDs.
    \end{itemize}

    \textbf{Output Template:}
\begin{verbatim}
{
  "triplets": [
    {   "subject": "{FRAME_ID}", 
        "predicate": "{RELATION}", 
        "object": "{FRAME_ID}" 
    }
  ]
}
\end{verbatim}

    \textbf{Rules:}
    \begin{itemize}
        \item (Frame ID Strictness) You must \textbf{ONLY} use the specific Frame IDs provided. If a frame is \texttt{None}, do not create a relationship to it.
        \item (Directionality) Ensure logical flow.
        \begin{itemize}
            \item Example: "North from X to Y" $\rightarrow$ \texttt{Y is\_north\_of X}.
            \item Example: "Kitchen next to Hall" $\rightarrow$ \texttt{Hall is\_next\_to Kitchen}.
        \end{itemize}
        \item (Ambiguity) If the directive cannot be resolved to the provided frames or is conversational filler, return an empty list \texttt{[]}.
    \end{itemize}

    \textbf{[Examples]}
\end{various}
\caption{Condensed prompt for the Linker module. The full prompt includes specific constraints on frame ID usage and directionality rules for resolving spatial relations to guide the model's reasoning process.}
\label{prompt:linker}
\end{figure}


\begin{figure}
\begin{reasoner}{Reasoner}
    \footnotesize
    \textbf{1. Master Planner}
    \begin{itemize}
        \item \textbf{Role:} Break down user questions into a high-level strategic plan.
        \item \textbf{Commands:}
        \begin{itemize}
            \item \texttt{POV}: Identify perspective (User A, B, or BOTH).
            \item \texttt{RAG[k=N]}: Retrieve top N relevant images/summary blocks.
            \item \texttt{PROCESS}: Filter, transform, or reason about gathered info.
            \item \texttt{FINAL\_ANSWER}: Formulate the response (Must be the last step).
        \end{itemize}
        \item \textbf{Output:} XML-like format \texttt{<answer><item> Step 1... </item>...</answer>}.
    \end{itemize}

    \vspace{0.3cm} \hrule \vspace{0.3cm}

    \textbf{2. Instruction Refiner}
    \begin{itemize}
        \item \textbf{Role:} Refine high-level plan steps into simple, direct natural language tasks for the Processor.
        \item \textbf{Rule:} Do not use SQL. Convert RAG instructions into search-friendly terms.
    \end{itemize}

    \vspace{0.3cm} \hrule \vspace{0.3cm}

    \textbf{3. Data Processor}
    \begin{itemize}
        \item \textbf{Role:} Execute specific instructions on provided context (e.g., "Find the image with the red bed").
        \item \textbf{Constraints:} Output \textbf{only} the requested result (raw \& clean). No conversational filler.
        \item \textbf{Logic:} Respect Temporal Logic (ImageID/BlockID = distinct events) vs Versioning (SequenceID = updates).
    \end{itemize}

    \vspace{0.3cm} \hrule \vspace{0.3cm}

    \textbf{4. Final Answerer}
    \begin{itemize}
        \item \textbf{Role:} Answer the specific sub-question using \textit{only} the retrieved info.
        \item \textbf{Format:}
\begin{verbatim}
<think> ... reasoning ... </think>
<answer> ... final verdict ... </answer>
\end{verbatim}
        \item \textbf{Logic:} If Yes/No question: return "yes"/"no" instead of IDs.
    \end{itemize}

    \textbf{[Examples]}
\end{reasoner}
\caption{Condensed prompt for the Reasoner module. The full prompt includes separate instructions for the Planner, Refiner, Processor, and Answerer agents, along with strict protocols for temporal logic and execution to guide the model's reasoning process.}
\label{prompt:reasoner}
\end{figure}


\begin{figure}[t]
\begin{judge}{Judge}
    \footnotesize
    You are a strict evaluator. Your task is to compare an LLM response against a correct response (ground truth) given a specific question and determine if they convey the same meaning. \\

    \textbf{[Contextualization]} \\

    \textbf{Inputs:}
    \begin{itemize}
        \item \textbf{Question:} The original query.
        \item \textbf{LLM Response:} The answer generated by the model.
        \item \textbf{Correct Response:} The ground truth answer.
    \end{itemize}

    \textbf{Output Template:}
\begin{verbatim}
<reasoning>
(1-2 sentences explaining why they are the same or different.)
</reasoning>
<answer>
(SAME or DIFFERENT)
</answer>
\end{verbatim}

    \textbf{Rules:}
    \begin{itemize}
        \item (Equivalence) Output \textbf{'SAME'} if the main content/meaning is identical.
        \begin{itemize}
            \item \textbf{Synonyms:} Count as same.
            \item \textbf{Style Differences:} "Yes, there was a car" vs "A car was there" $\rightarrow$ SAME.
            \item \textbf{Negation Styles:} "Do not recall a cat" vs "No cats were there" $\rightarrow$ SAME.
        \end{itemize}
        \item (Differentiation) Output \textbf{'DIFFERENT'} if meanings diverge.
        \begin{itemize}
            \item \textbf{Logical Negation:} If one contradicts the other.
            \item \textbf{Distinct Entities:} "Kitchen" vs "Dining Room" $\rightarrow$ DIFFERENT (do not reason about similar functions).
        \end{itemize}
        \item (Focus) Ignore intermediate reasoning; judge \textit{only} the final answer provided at the end.
    \end{itemize}

    \textbf{[Examples]}
\end{judge}
\caption{Condensed prompt for the Judge module. The full prompt includes specific criteria for distinguishing semantic meaning from stylistic differences and handling negations to guide the model's reasoning process during evaluation.}
\label{prompt:judge}
\end{figure}

\begin{figure}[t]
\newpage
\begin{various}{Annotator}
    \footnotesize
    You are an expert Data Annotation Specialist for a Conversational AI dataset.
    Your task is to analyze a specific Question-Answer pair from a dialogue about a physical environment and classify its complexity, constraints, and validity.
    \\

    \textbf{Inputs:}
    \begin{itemize}
        \item \textbf{Relation Type:} \texttt{\{\{ relation\_type \}\}}
        \item \textbf{Question:} \texttt{"\{\{ question \}\}"}
        \item \textbf{Gold Answer:} \texttt{"\{\{ gold\_answer \}\}"}
    \end{itemize}

    \textbf{Classification Definitions \& Rules:}
    \begin{itemize}
        \item (Complexity Type) \textbf{Local vs. Relational:} Determine if the question can be answered by looking at the content of a single room visualized in isolation, or if it requires traversing a map/timeline.
        \begin{itemize}
            \item \textbf{Local:} The target is identified by its internal visual attributes or specific name. If you were teleported directly into the target room, you could answer the question without knowing how you got there or when you visited it. (Examples: "What color is the wall in the kitchen?", "Which room has a red couch?", "Is there a TV in this room?")
            \item \textbf{Relational:} The target is identified by its spatial connection to other rooms or its order in time. You need a map or a history log to identify the correct target. (Examples: "What is in the room north of the kitchen?", "What was the first room I visited?", "Is the room I am in now bigger than the previous one?")
        \end{itemize}
        \item (Question Type) \textbf{Binary vs. Open:}
        \begin{itemize}
            \item \textbf{Binary:} The question demands a Yes/No/True/False verification.
            \item \textbf{Open:} The question asks for specific content, counts, or descriptions.
        \end{itemize}
        \item (Constraint Type) \textbf{List vs. Free:}
        \begin{itemize}
            \item \textbf{List:} The question explicitly lists candidates to check (e.g., "Which among the kitchen, bath, or hall...?", "Is it red or blue?").
            \item \textbf{Free:} The search space is open to the entire context or vocabulary.
        \end{itemize}
        \item (Validity Type) \textbf{Valid vs. Missing:}
        \begin{itemize}
            \item \textbf{Valid:} The answer provides the requested information or definitively confirms/denies a fact.
            \item \textbf{Missing:} The answer states that the premise of the question is false, the object was not mentioned, or the information is unknown.
        \end{itemize}
    \end{itemize}
    
    \textbf{Output Format Rules} \\
    Return ONLY a single valid JSON object. The keys must exactly match the provided output template.
    JSON Template:

\begin{verbatim}
{ "complexity_type": "local", "question_type": "binary", 
"constraint_type": "free", "validity_type": "valid" }
\end{verbatim}

\end{various}
\caption{Full prompt for the Annotator, used for the quantitative analysis. `Validity' was never used in the work.}
\label{prompt:annotator}
\end{figure}


\clearpage



\end{document}